\pdfoutput=1

\documentclass[11pt]{article}

\usepackage[]{acl}  
\usepackage{nert}

\usepackage{times}
\usepackage{latexsym}

\usepackage[utf8]{inputenc}

\usepackage[
singlelinecheck=false 
]{caption}
\usepackage{xcolor}
\usepackage{booktabs,colortbl, array}
\usepackage{pgfplotstable}
\pgfplotsset{compat=1.8}

\definecolor{rulecolor}{RGB}{0,71,171}
\definecolor{tableheadcolor}{gray}{0.92}

\newcommand{\topline}{ %
        \arrayrulecolor{rulecolor}\specialrule{0.1em}{\abovetopsep}{0pt}%
        \arrayrulecolor{tableheadcolor}\specialrule{\belowrulesep}{0pt}{0pt}%
        \arrayrulecolor{rulecolor}}
\newcommand{\midtopline}{ %
        \arrayrulecolor{tableheadcolor}\specialrule{\aboverulesep}{0pt}{0pt}%
        \arrayrulecolor{rulecolor}\specialrule{\lightrulewidth}{0pt}{0pt}%
        \arrayrulecolor{white}\specialrule{\belowrulesep}{0pt}{0pt}%
        \arrayrulecolor{rulecolor}}
\newcommand{\bottomline}{ %
        \arrayrulecolor{white}\specialrule{\aboverulesep}{0pt}{0pt}%
        \arrayrulecolor{rulecolor} %
        \specialrule{\heavyrulewidth}{0pt}{\belowbottomsep}}%

\newcommand{\midheader}[2]{%
        \midrule\topmidheader{#1}{#2}}
\newcommand\topmidheader[2]{\multicolumn{#1}{c}{\textsc{#2}}\\%
                \addlinespace[0.5ex]}

\pgfplotstableset{normal/.style ={%
        header=true,
        string type,
        column type=l,
        every odd row/.style={
            before row=
        },
        every head row/.style={
            before row={\topline\rowcolor{tableheadcolor}},
            after row={\midtopline}
        },
        every last row/.style={
            after row=\bottomline
        },
        col sep=&,
        row sep=\\
    }
}

\title{A Multi-Encoder Frozen-Decoder Approach for Fine-Tuning Large Language Models\\[15pt]
}

\author{Kaustubh D. Dhole  \\
    Department of Computer Science \\
  Emory University \\
  Atlanta, USA \\
  \eml{kdhole@emory.edu} \\
}

\date{}

\begin{document}
\maketitle
\begin{abstract}
Among parameter-efficient fine-tuning methods, freezing has emerged as a popular strategy for speeding up training, reducing catastrophic forgetting, and improving downstream performance. We investigate the impact of freezing the decoder in a multi-task setup comprising diverse natural language tasks, aiming to reduce deployment overhead and enhance portability to novel tasks. Our experiments, conducted by fine-tuning both individual and multi-task setups on the AlexaTM model, reveal that freezing decoders is highly effective for tasks with natural language outputs and mitigates catastrophic forgetting in multilingual tasks. However, we find that pairing frozen decoders with a larger model can effectively maintain or even enhance performance in structured and QA tasks, making it a viable strategy for a broader range of task types.
\end{abstract}

\section{Introduction}

Language models utilized in multi-task setups have long been shown to have improved performance across a variety of natural language tasks~\cite{10.1145/1390156.1390177}. While training multiple tasks together lets individual tasks benefit from each others' data, a single model trained on multiple tasks offers lesser control of individual capabilities. This parameter-sharing paradigm is unarguably less computationally efficient as it necessitates retraining the complete model on the introduction of a new task. 

On the other hand, fine-tuning individual task-specific models offers increased control and modularity at the cost of lesser training data. Our work attempts to further exploit this latter paradigm in a virtual assistant experimentation platform to check for performance benefits via partial fine-tuning. We explore partial fine-tuning in the context of task-specific encoder-decoder models. 

\begin{figure}
    \includegraphics[width=\columnwidth]{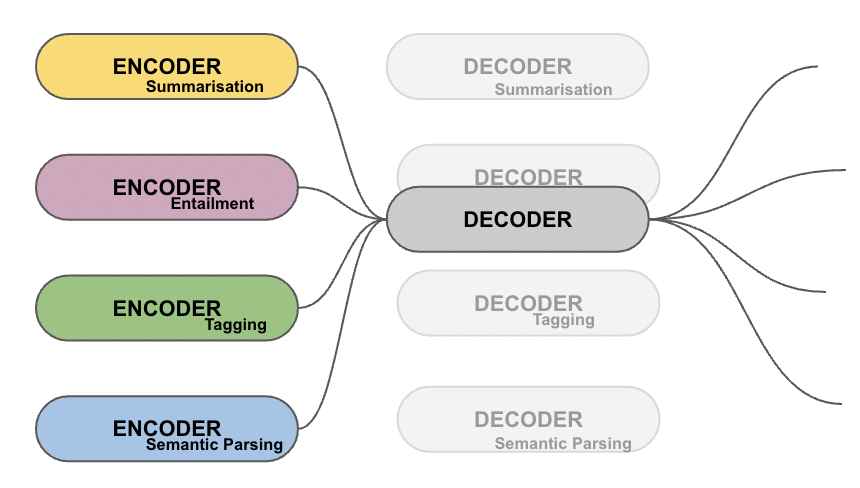}
    \caption{Freezing the decoder helps train faster, reduces deployment overhead, improves portability, and possibly increases task performance. }
    \label{fig:multi_encoder}
\end{figure}

Among partial and parameter-efficient fine-tuning methods~\cite{han2024parameterefficient}, freezing has been a popular strategy adopted to train faster, reduce catastrophic forgetting, and increase downstream performance. We seek to investigate the benefits and drawbacks of specifically freezing the parameters of the decoder while fine-tuning individual task-specific encoders (Figure~\ref{fig:multi_encoder}).

This delineation of training of the encoder and the decoder permits us to look at both of them separately. Freezing the parameters of the decoder has shown to be effective for improving downstream performance on machine translation~\cite{cooper-stickland-etal-2021-recipes}. If a sub-module of the network is well pre-trained and exposed to data from languages or domains similar to the fine-tuning task, freezing it might not necessarily restrict its learning capability and may even enhance it. Eg.~\citet{cooper-stickland-etal-2021-recipes} observe that fine-tuning mBART~\cite{liu-etal-2020-multilingual-denoising} with a frozen decoder performs better than mBART with all trainable layers, especially if the translating output language is English. When the translating source language is English, freezing the encoder helps gain performance. 

In general, freezing reduces the overall number of parameters to train improving training efficiency. Particularly in the case of mBART, freezing the decoder can help almost double the number of tokens that could be fed in while training as the memory required to store decoder gradients can be freed up~\cite{cooper-stickland-etal-2021-recipes}. More importantly, frozen decoders can help reduce deployment overhead in ``batched inference'' settings by reducing the number of employed decoders.

It can hence be imperative to understand if any performance gains can be exploited via deploying a common decoder for natural language tasks. In this work, we study the impact of different decoding setups against different natural language processing (NLP) tasks spanning various task types.  Our contributions are the following:
\begin{itemize}
    \item We first investigate the downstream performance impact of freezing the decoder
    \item Further, if there is a drop in performance, we attempt to mitigate it by increasing the decoder's size while still keeping it frozen
    \item Finally, we also perform a comparison against a single model trained on all tasks along with a frozen decoder comparison of the same.
\end{itemize}

\section{Related Work}
Parameter freezing has always been popular albeit with different objectives eg. Dropout~\cite{https://doi.org/10.48550/arxiv.1207.0580} \& Stochastic Depth (dropping of ResNet blocks while training)~\cite{10.1007/978-3-319-46493-0_39}. In encoder-decoder models, freezing has been explored a bit for machine translation.~\citet{thompson-etal-2018-freezing, zoph-etal-2016-transfer} explore different ways to freeze parameters of RNNs.~\citet{cooper-stickland-etal-2021-recipes} find that freezing encoder or decoder parameters improve performance on distantly related, language directions.~\citet{brock2017freezeout} presents a technique, FreezeOut by progressively freezing hidden layers of popular computer vision models.~\citet{lee2019elsa} perform experiments on freezing the final layers of BERT~\cite{kenton2019bert} \& RoBERTa~\cite{liu2019roberta} and observe that 90\% of their performance can be retained by only fine-tuning 25\% of the final layers and that freezing a few layers increases performance in a few tasks like SST-2~\cite{socher2013recursive}.~\citet{he-etal-2021-effectiveness} show that parameter efficient adapter based fine-tuning of BERT \& RoBERTa outperforms traditional fine-tuning on low-resource and cross-lingual tasks and mitigates catastrophic forgetting.

\section{Experiments}
We perform our experiments on two specific encoder-decoder models, a.k.a. Alexa Teacher Model (AlexaTM) released by ~\citet{Soltan2022, FitzGerald2022}. 

\subsection{Models}
\textbf{AlexaTM}: In all our experiments, we use the large version of AlexaTM which is pre-trained similar to mBART Large~\cite{liu-etal-2020-multilingual-denoising} with 12-layer encoders and 12-layer decoders. This 511 million parameter model has been exposed to spoken and written formats of Arabic, English, French, German, Hindi, Italian, Japanese, Marathi, Portuguese, Spanish, Tamil and Telugu. \\\\\textbf{AlexaTM 2B}: Since we would like to address possible performance drops, we perform the same experiments with the larger version (2 billion parameters) of the same model. We keep the decoder here frozen too.

\subsection{Tasks}
We perform our experiments on multiple NLP datasets spanning tasks of diverse types.
\begin{itemize}
    \item \textbf{XSUM} \cite{Narayan2018DontGM} A real-world, large-scale summarization dataset collected from BBC to generate one-sentence news summaries.
    \item \textbf{MTOP} \cite{li-etal-2021-mtop} A task-oriented semantic parsing dataset covering 6 languages and 11 domains.
    \item \textbf{SQUAD v2} \cite{rajpurkar-etal-2018-know} A large scale reading comprehension dataset with unanswerable queries.
    \item \textbf{MASSIVE} \cite{bastianelli-etal-2020-slurp,fitzgerald2022massive} About 1M labeled utterances spanning 51 languages for intent classification and slot tagging.
    \item \textbf{XNLI} \cite{conneau2018xnli} Cross-lingual natural language inference in 15 languages
    \item \textbf{CommonGen} \cite{lin-etal-2020-commongen} Generating a sentence using a given set of concept words that describe everyday life scenarios.
    \item \textbf{WebNLG} \cite{zhou-lampouras-2020-webnlg} RDF-to-Text Generation.
\end{itemize}

\subsection{Training \& Evaluation}
Each dataset is sourced from HuggingFace~\cite{wolf-etal-2020-transformers} or the corresponding publicly available repository and is used to fine-tune 3 instances of encoder-decoders 1) AlexaTM with a trainable decoder, 2) AlexaTM with a frozen decoder and 3) AlexaTM 3B with a frozen decoder. Models are trained in batches of 128 on EC2 P3 Instances (8 NVIDIA® V100 Tensor Core GPUs). Adam Optimization~\cite{kingma2014adam} with a learning rate of 10\textsuperscript{6} and a linear decay to $lr = 5x10\textsuperscript{6}$ over 100k updates is employed. For high-throughput training, Deepspeed~\cite{deepspeed, zeroRajBhandari} is used. It permits partitioning model parameters, optimizer states, and gradients across the 8 GPUs. Model checkpoints were saved either based on the perplexity or exact match scores over the validation set.

Additionally, we fine-tune the aforementioned three models using a mixture of XSUM, XNLI, CommonGen, WebNLG, and MTOP. Each dataset is sampled proportional to the square root of the number of examples in that dataset.

\begin{table*}[h]
    \centering
    \begin{tabular}{lccc}
        \toprule
        Dataset & Original Size & Sampling Size & Proportion (\%) \\
        \midrule
        WebNLG    & 35k  & 188  & 9.34  \\
        CommonGen & 67k  & 260  & 12.92 \\
        XSUM      & 204k & 452  & 22.45 \\
        XNLI      & 393k & 627  & 31.15 \\
        MTOP      & 16k  & 125  & 6.21  \\
        SQUAD     & 130k & 361  & 17.93 \\
        \bottomrule
 \end{tabular}
    \captionsetup{justification=centering}
    \caption{Proportion of Each Dataset}
    \label{tab:dataset_proportion}
\end{table*}

For the evaluation of XSUM, CommonGen, and WebNLG tasks, we utilize NLG metrics from GEM-metrics~\cite{gehrmann-etal-2021-gem, gehrmann2022gemv2}. For SQUADv2, we use the official evaluation script~\cite{rajpurkar-etal-2018-know} and report exact match scores. For MTOP, XNLI, and MASSIVE, we simply perform an exact match.

\section{Results \& Analysis}

\subsection{NLG Tasks}
Table~\ref{nlgresults} displays the results on NLG tasks. 3 scores are computed -- the ROGUE scores~\cite{lin2004rouge}, the BLEU~\cite{papineni2002bleu} metric and the NIST metric~\cite{doddington2002automatic}. The ROGUE metrics measure the recall of words and n-grams off the reference summaries and BLEU measures the precision. The NIST scores are just like BLEU scores but they give more emphasis to rare words and phrases. 

For XSUM, freezing the decoder of AlexaTM has just close to a 2\% decrease in the ROUGE scores, and performance is improved with the larger frozen decoder. When fine-tuned on WebNLG, freezing results in a 2\% drop in performance, and the performance is retained with the larger frozen decoder. So, this is interesting since freezing improves the performance in the single task setting. It might not be surprising considering that a task like WebNLG has properties similar to the pre-training data format. Besides, mixing the tasks helps WebNLG drastically but not XSUM. Interestingly, XSUM’s data is beneficial for XSUM as well as WEBNLG. But it is also likely that XSUM's double share of training data might be a reason. 

For WebNLG, the mixed model setup was better than the single model setup, except for the large frozen decoder setting which showed the best performance. The 2B model seemingly benefited more from the large pre-training data it was exposed to, as compared to cross-task data during mix training. \textbf{The cross-task data is indeed useful but if the decoder is well-pre-trained the task doesn't need cross-task data.} Freezing greatly benefits tasks that have properties similar to the pre-training data format. The performance of WebNLG, remains almost the same on usage of a frozen decoder unlike MTOP since WebNLG has natural language target sequences.

\begin{table*}[h]
    \centering
    \begin{tabular}{lccccc}
        \toprule
        Task & ROUGE-1 & ROUGE-2 & ROUGE-L & BLEU & NIST \\
        \midrule
        \multicolumn{6}{c}{Summarization (XSUM)} \\
        AlexaTM                & .39 & .17 & .32 & 11.78 & 4.39 \\
        AlexaTM (frozen)       & .37 & .15 & .30 & 9.87  & 4.05 \\
        AlexaTM-2B (frozen)    & .39 & .16 & .31 & 10.84 & 4.27 \\
        \hline
        AlexaTM-mix            & .39 & .15 & .31 & 9.50  & 3.92 \\
        AlexaTM-mix (frozen)   & .40 & .17 & .32 & 11.37 & 4.42 \\
        \midrule
        \multicolumn{6}{c}{Data-To-Text (WebNLG)} \\
        AlexaTM                & .67 & .41 & .51 & 29.76 & 6.49 \\
        AlexaTM (frozen)       & .69 & .43 & .54 & 32.01 & 6.98 \\
        AlexaTM-2B (frozen)    & .74 & .47 & .58 & 35.14 & 7.59 \\
        \hline
        AlexaTM-mix            & .72 & .45 & .55 & 32.53 & 7.30 \\
        AlexaTM-mix (frozen)   & .70 & .44 & .55 & 30.53 & 6.68 \\
        \midrule
        \multicolumn{6}{c}{Data-To-Text (CommonGen)} \\
        AlexaTM                & .59 & .21 & .50 & 13.12 & 4.26 \\
        AlexaTM (frozen)       & .59 & .21 & .50 & 13.14 & 4.18 \\
        AlexaTM-2B (frozen)    & .54 & .21 & .44 & 11.00 & 4.05 \\
        \hline
        AlexaTM-mix            & .55 & .21 & .44 & 11.02 & 4.17 \\
        AlexaTM-mix (frozen)   & .56 & .22 & .46 & 10.95 & 3.87 \\
        \bottomrule
    \end{tabular}
    \captionsetup{justification=centering}
    \caption{Performance of Frozen and Trainable Decoders on Generation Tasks with num\_beams=3.}
    \label{nlgresults}
\end{table*}

It is in the case of MTOP where we notice the biggest drop in performance on using a frozen decoder of the same size. One could perform extensive hyperparameter tuning eg. evaluate with more beams, early stopping, etc. but with num\_beams=3, we see almost a 14\% drop on using a frozen decoder. This doesn’t look so surprising if we look at the format of the output sequences which are unlike spoken or written text i.e. they radically differ from the format of the pre-training data. Intriguingly, with a larger version of the decoder, freezing makes up for quite a bit of the performance loss as well as a 2\% increase. And in the mixed setting, MTOP performs poorly possibly due to the domination of other text output datasets.

\begin{table*}[h]
    \centering
    \begin{tabular}{lcc}
        \toprule
        Setup & Exact Match (MTOP) & Exact Match (\% SQuAD) \\
        \midrule
        \multicolumn{3}{c}{Semantic Parsing (MTOP) and QA (SQuAD)} \\
        AlexaTM                & .66  & 72.33 \\
        AlexaTM (frozen)       & .52  & 69.96 \\
        AlexaTM-2B (frozen)    & .68  & 74.15 \\
        \hline
        AlexaTM-mix            & .27  & 61.12 \\
        AlexaTM-mix (frozen)   & .28  & 62.51 \\
        \bottomrule
    \end{tabular}
    \captionsetup{justification=centering}
    \caption{Performance on Structured Generation (MTOP) and QA (SQuAD) with num\_beams=3.}
    \label{combined_results}
\end{table*}
In XNLI, mixing the tasks benefits the model in both the frozen as well as the trainable setting. Freezing the parameters improves performance. However, we also noticed that by increasing the number of beams from 3 to 5, 10 and 20, the gap between these setups tends to decrease.

Question Answering over SQUAD is best with the frozen version of the larger decoder. Mixing the tasks hurts performance for both trainable as well as frozen settings.

For MASSIVE, we first train only on the English intents dataset of MASSIVE and then evaluate all languages. While the MASSIVE dataset has intents for 59 languages, we report performance only for those languages on which AlexaTM was pre-trained. These are the most interesting results in this set of experiments. We see extremely substantial increases in performances on freezing the decoder for every language except for English - the language on which the model was fine-tuned where there is an increase of 4\%. On freezing, there is an 8\% increase for Spanish, 9\% for Arabic, and 10\% for Italian. \textbf{This strongly suggests how catastrophic forgetting can be really damaging to a model when trained on a specific task or language’s data (English in this case) and how powerful freezing can be as a remedy.} So, decoder freezing seems to be a powerful remedy to not only improve absolute performance but also avoid catastrophic forgetting. More importantly, with a larger frozen decoder, we get a substantial boost in performance across all languages. 

We also perform a separate set of experiments to let intents and slots be jointly trained. Instead of training two separate parameters, we make the model predict the single flat semantic parse which contains the intent as well as the slots. We notice that joint training helps performance for intent classification. Even in this case, frozen decoders help address catastrophic forgetting and improve performance across all languages.

\section{Conclusion}

Our results show that freezing decoders can lead to significant efficiency gains and mitigate catastrophic forgetting, especially in multilingual and classification tasks. Freezing is particularly effective for tasks with natural language target sequences, such as WebNLG and CommonGen, where performance remains comparable or improves slightly. For structured tasks like MTOP, freezing leads to performance degradation unless paired with a larger decoder.

Freezing decoders in multilingual tasks, such as MASSIVE, yields substantial performance improvements by preventing catastrophic forgetting. This is particularly evident in non-English languages, where freezing improves performance by 8\% to 10\%. Joint training of intents and slots further enhances performance, reinforcing the role of freezing as a regularizer.

While freezing decoders shows promise for efficient fine-tuning, mixed-task setups often face cross-task interference, especially in structured and QA tasks. Future work could explore more robust strategies for mitigating cross-task interference and evaluate the impact of freezing other parts of the network.

A production system could benefit from deploying frozen decoders for groups of tasks with common output formats, reducing deployment overhead and improving modularity. This approach can enable efficient task fine-tuning by only updating encoder weights while keeping decoders fixed.

Overall, our findings suggest that freezing decoders can be a powerful approach for parameter-efficient fine-tuning, particularly when tasks with similar output formats are grouped together.

\section{Future Work}
A vital evaluation from the perspective of multi-task design would be to also perform a parallel evaluation with freezing encoders and other parameters of the whole network partially. Future work would explore the same. Performances on mix tasks can be highly variable being dictated by the dataset proportion and the choice of other datasets. It would be interesting to explore the downstream effects of different taxonomies~\cite{fifty2021efficiently} of a larger number of tasks~\cite{srivastava2023beyond}, as well as resort to data augmentation~\cite{dhole-etal-2023-nl,mille1automatic} for up-sizing smaller datasets. 

\section*{Ethical Considerations}
Large language models (LLMs) should always be examined from a sociotechnical perspective~\cite{dhole2023large}. Appropriate guardrails must be employed, as deployment contexts may differ from the task-specific evaluations that we conducted. Thorough testing across diverse scenarios is crucial to ensure robustness, mitigate biases, and prevent potential harm or unintended consequences.

\begin{table*}
        \centering
        \pgfplotstabletypeset[normal,
                columns/eg/.style={
                column name={$E_{\textup{g}}$ (\si{\electronvolt})},
                dec sep align
        }
        ]{ %
        Setup & es & ar & it & en & de & fr & ja & te & hi\\
        \topmidheader{5}{Intent Classification (MASSIVE)}
        AlexaTM & 52.49 & 35.84 & 50.87 & 79.32 & 55.61 & 53.4 & 33.52 & 27.24 & 37.12 \\
        AlexaTM (frozen) & 61.06 & 44.99 & 60.69 & 83.12 & 64.06 & 64.49 & 41.93 & 36.62 & 48.02\\
        AlexaTM-2B (frozen) & 77.47 & 62.91 & 78.58 & 87.96 & 78.95 & 78.17 & 73.24 & 65.53 & 73.13\\
        \midheader{5}{Intent Classification (Joint) (MASSIVE)}
        AlexaTM & 55.48 & 38.77 & 53.29 & 82.25 & 59.11 & 57.06 & 35.64 & 29.32 & 43.51\\
        AlexaTM (frozen) & 65.00 & 46.03 & 63.28 & 83.65 & 66.27 & 69.57 & 41.56 & 39.87 & 50.6\\
        AlexaTM-2B (frozen) & 78.51 & 64.02 & 78.92 & 87.86 & 79.69 & 81.03 & 70.81 & 63.79 & 72.63\\
        }
        \caption{Performance of Frozen and Trainable Decoders on Dialogue Tasks}
\end{table*}

\begin{table*}[h]
    \centering
    \begin{tabular}{lrr}
        \toprule
        Setup & Accuracy (3) (\%) & Accuracy (20) (\%) \\
        \midrule
        \multicolumn{3}{c}{Entailment (XNLI)} \\
        AlexaTM                & 51.73 & 73.37 \\
        AlexaTM (frozen)       & 65.50 & 84.21 \\
        AlexaTM-2B (frozen)    & 83.67 & --    \\
        \hline
        AlexaTM-mix            & 84.29 & --    \\
        AlexaTM-mix (frozen)   & 83.91 & --    \\
        \bottomrule
    \end{tabular}
    \captionsetup{justification=centering}
    \caption{Performance of Frozen and Trainable Decoders on Classification Tasks.}
    \label{classification_results}
\end{table*}

\bibliography{mypaper}

\end{document}